\documentclass[conference]{IEEEtran}
\IEEEoverridecommandlockouts
\usepackage{cite}
\usepackage{amsmath,amssymb,amsfonts}
\usepackage{algorithmic}
\usepackage{graphicx}
\usepackage{textcomp}
\usepackage{xcolor}
\def\BibTeX{{\rm B\kern-.05em{\sc i\kern-.025em b}\kern-.08em
    T\kern-.1667em\lower.7ex\hbox{E}\kern-.125emX}}
    \usepackage{changes}
\begin{document}

\title{A blindspot of AI ethics: anti-fragility in statistical prediction 
}

\author{\IEEEauthorblockN{1\textsuperscript{st} Michele Loi}
\IEEEauthorblockA{\textit{Digital Society Initiative} \\
\textit{University of Zurich}\\
Zurich, Switzerland \\
michele.loi@uzh.ch}
\and
\IEEEauthorblockN{2\textsuperscript{nd} Lonneke van der Plas}
\IEEEauthorblockA{\textit{Institute of Linguistics and Language Technology} \\
\textit{University of Malta}\\
Msida, Malta \\
Lonneke.vanderPlas@um.edu.mt}
}

\maketitle

\begin{abstract}
With this paper, we aim to put an issue on the agenda of AI ethics that in our view is overlooked in the current discourse. The current discussions are dominated by topics such as trustworthiness and bias, whereas the issue we like to focus on is counter to the debate on trustworthiness. We fear that the overuse of currently dominant AI systems that are driven by short-term objectives and optimized for avoiding error leads to a society that loses its diversity and flexibility needed for true progress. We couch our concerns in the discourse around the term anti-fragility and show with some examples what threats current methods used for decision making pose for society.  
\end{abstract}


\section{Anti-fragility and AI}

The term anti-fragility is used by Nassim Taleb to describe things that gain from disorder \cite{Taleb:Antifragility:2012}. More precisely, fragility is related to how a system suffers from the variability of its environment beyond a certain preset threshold, while anti-fragility is the opposite of this, namely it refers to when systems benefit from this variability \cite{TalebDouady:Mathematical:2013}.

Crucially, anti-fragility is not the same as robustness. A robust object or system is left relatively unaffected by any extreme variability of its environment (in the sense that it will tend to recover its initial state). By contrast, an anti-fragile one will be affected by the same variability, but positively. In machine learning, the robustness of a system refers to its ability to cope with independent test sets and errors in training data. Our argument, familiar to readers of Taleb's work, is that decisions that are based on statistical predictions that maximize utility (or equivalently, reduce some kind of error or loss) are predictably subject to a specific type of failure, connected to anti-fragility. Our contention is, roughly, that when a certain type (more in particular, an optimizing kind) of decision becomes too widespread, and prevails across domains, it makes society lose its anti-fragility. And this is a really bad thing.

The problem we are concerned with is also one of scale: it would emerge if most organizations are run following the recommendations and decisions stemming from statistical models that learn from past data. The problem would be aggravated if some regulatory authorities were also run on the basis of a similar logic. The risk is that one misses out on the diversity and variability of forms of organization that appear less efficient, at least on paper. Consider Taleb's concept of a "Black Swan": a low probability event with a high impact\cite{Taleb:Black:2007}. Unplanned diversity plays a key role in the recovery after a Black Swan. It often leads the discovery of new, entirely unexpected, opportunities. Here we focus on Black Swans that generate unexpected social benefits and we draw attention to the risk for society to lose its potential to gain from low probability events with a high impact. This will happen if the commercially dominant type of AI - decision systems, that make predictions on the basis of models trained on available data - becomes too widespread,  minimizing short-time loss at the expense of the benefits more chaotic, error-prone, and inefficient working of human systems.
 
We fear that, in spite of the considerable attention devoted to the ethics of AI over the last few years, this problem is currently off the maps of the debate on AI ethics.

\section{The social good of anti-fragility}

Taleb's concern with anti-fragility has an interesting pedigree in (at least) Western political philosophy. It is deeply related to a vision of human affairs that the economist and historian of ideas Thomas Sowell has labelled the "constrained vision"  \cite{Sowell:Conflict:2007,Murphy:Unconstrained:2015}. Thinkers like Adam Smith, Edmund Burke, and Friedrich Hayek all share the same view of humans as fundamentally constrained in both their moral and intellectual capacities. They are highly skeptical of theoretical knowledge and value the implicit knowledge, embodied in traditions, which reflects the inarticulate experiences of the many, filtered by history. They have provided the intellectual arguments for institutions that achieve the best results when many people, even most people, do the morally wrong or intellectually uninformed thing, and when there is no decent theory to guide them. 

One example of this is the restaurants market, where most shops that open are likely to be soon out of business. And yet most individuals have (irrational) faith that they will be successful. As Taleb (2012) \cite{Taleb:Antifragility:2012} points out "[n]atural and nature like systems want some overconfidence on the part of individual economic agents, i.e. [...] the underestimation of the risk of failure in their businesses, provided that their failures does not impact others [...]" (p.75). Another example is evolution through natural selection, which relies on random mutations, which more often than not end up being harmful for the individual affected, while enabling adaptation to the widest range of natural environments. 

Biological life and the restaurant market do not merely tolerate imprudence. They are anti-fragile: they thrive from individual mistakes and random variation. While the likelihood for an individual restaurant to go out of business is high (which implies wasted resources), the restaurant market improves steadily, retaining its plasticity and generating novelties. Mistakes harm individuals but benefit society, by revealing opportunities that are otherwise ignored. Such nature-like systems are inefficient in the short term, but utilitarian in the long term, since the whole benefits from what harms some, sometimes most, individuals. Compared to systems based on predictions that favor the likely-to-be-successful, they enable flexibility by promoting the odd one out. The restaurant market is robust: it may adapt to a change in available ingredients, culinary preferences, and the depth of clients' pockets. It is also anti-fragile: it is able to gain from such shock, generating new, tasteful, exciting types of foods. 

\section{Anti-fragility and AI: some examples}

We shall now present some hypothetical cases in which the use of statistical-prediction based decision rules may intuitively reduce the anti-fragility of (A) the life of an individual, (B) science (C) the economy. 

A) How to fail to stumble upon the love of your life: You regularly commute to work, walking your way from the train station to your office. There are several possible routes you may take. You open Google Maps and choose the shortest route, and you are stuck to that path every day. Nothing interesting ever happens as you quickly walk between work and your train. You are single. Though you tried several dating websites and apps, you still have not found the life companion you were looking for. If you had only decided to stop allowing Google Maps to tell you what to do, if you had started to take longer, more random paths, exploring the surroundings, you might have noticed, one day, a music school, and decided to take guitar lessons. Besides learning what is now your favorite hobby, you would eventually have met your future wife at the local guitar shop.  

B) The non-discovery of penicillin: You are Alexander Fleming, a Scottish researcher, in a hypothetical alternative past in which you decide on your daily experiment by following the recommendations of a data-driven system. As you return from your summer holiday, you do not realize that something strange has happened to the bacterial cultures stacked on a bench in the corner of your laboratory. You are too eager to read what awaits you on your data-driven research recommendation system. In your urge to improve your academic CV, you only pursue the research question predicted to generate the paper most likely to be published in a high-impact journal. The model which you rely on uses a similarity metric, based on past publications, and has proven reliable in the past. So you ask your Ph.D. student to tidy up and you do not waste a single moment of your precious time looking at what has happened on the bench.

C) How Steve Jobs failed to get his first Silicon Valley job: You are the manager of Atari in an alternative past in which algorithmic job-aptitude scoring have been invented before the PC and the I-Phone. You have just interviewed a young man called Steve Jobs. You are impressed by his  enthusiasm and vision in the field of video-games, but you are a rational, data-driven person who does not act based on gut feelings. The software you use (trained on the personality traits of past high-achievers), ranks him as the employee with the lowest potential you have ever interviewed. So you say no to Steve Jobs. Not seeing a future in the software industry, the young Steve Jobs moves permanently to India, where he becomes the most popular spiritual guru of the Eighties after Osho.

All examples provided describe different versions of the same phenomenon. In these dystopian alternate universes, we experience a loss of social utility (in case A, individual utility) deriving from anti-fragile systems. They all describe the same pattern, where the maximization of some narrowly defined utility function with the help of statistical generalizations from the past, enhances short-term efficiency for the individual at the expense of the capacity, for the individual (in some cases) and the whole of society, to benefit from unknown unknowns. 

Our point here is not that AI will necessarily drive us into such future: this is not an argument against AI technology as such. Instead, there are approaches to machine learning which are more favorable to exploration and less to immediate exploitation of the results. The risk does not lie in the technology, but in the incentive system generated by short-sighted market forces combined with an excessive ethical focus on harm prevention. Ultimately, the question here is ethical and societal. The good of anti-fragility has to do with the forgone potential gains from accepting what seems socially undesirable, namely a less predictable future and a higher risk of failure. When the focus is on robustness, but anti-fragility gets forgotten, the focus becomes preventing mistakes due to overfitted models. Calls for AI to be more robust are often reductively interpreted as calls to reduce the error rate (due to overfitting), when moving from training to independent test data. Our focus here is more holistic and long-term. We are suggesting that some error with regards to short-term objectives is the price we have to pay to achieve long term anti-fragility in society, and with this paper, we hope to put it on the agenda of AI ethics.

\bibliographystyle{./IEEEtran}
\bibliography{./IEEEabrv,./IEEEexample,AI_antifragility}

\end{document}